\documentclass[journal,comsoc]{IEEEtran}
\usepackage[T1]{fontenc}
\usepackage{xspace}
\usepackage{enumitem}
\usepackage{epsfig}
\usepackage{graphicx}
\usepackage{amsmath}
\usepackage{amsfonts}
\usepackage{amssymb}
\usepackage{array}
\usepackage{rotating}
\usepackage{epstopdf}
\usepackage{caption}
\usepackage{xcolor}
\usepackage{subfig}
\usepackage{gensymb}
\usepackage{multirow}
\usepackage{sansmath}
\usepackage[hyphens]{url}
\usepackage[hidelinks]{hyperref}
\usepackage{enumitem}
\usepackage[autostyle]{csquotes}
   \MakeAutoQuote{‘}{’}

\ifCLASSINFOpdf
\else
\fi
\usepackage{amsmath}
\usepackage[cmintegrals]{newtxmath}
\def\eg{\textit{e.g.}\@\xspace} 
\def\ie{\textit{i.e.}\@\xspace}

\def\etc{\textit{etc}.\@\xspace} 
\def\wrt{w.r.t.\@\xspace} 

\begin{document}
%
\title{Demographic Fairness in Biometric Systems: \\What do the Experts say?}
%
%
%

\author{Christian Rathgeb,~Pawel Drozdowski,~Naser Damer, Dinusha C. Frings, Christoph Busch
\thanks{C. Rathgeb, P. Drozdowski, and C. Busch are with Hochschule Darmstadt, Germany.}
\thanks{N. Damer is with Fraunhofer Institute for Computer Graphics Research and the TU Darmstadt, Darmstadt, Germany.}
\thanks{D. C. Frings is with the European Association for Biometrics.}}

%
%

\markboth{}%
{}
%



\maketitle

\begin{abstract}
Algorithmic decision systems have frequently been labelled as ``biased'', ``racist'', ``sexist'', or ``unfair'' by numerous media outlets, organisations, and researchers. There is an ongoing debate whether such assessments are justified and whether citizens and policymakers should be concerned. These and other related matters have recently become a hot topic in the context of biometric technologies, which are ubiquitous in personal, commercial, and governmental applications. Biometrics represent an essential component of many surveillance, access control, and operational identity management systems, thus directly or indirectly affecting billions of people all around the world. In order to provide a forum for experts in the field, the European Association for Biometrics organised an event series with ``demographic fairness in biometric systems'' as an overarching theme. The events featured presentations by international experts from academic, industry, and governmental organisations and facilitated interactions and discussions between the experts and the audience.  Further consultation of experts was undertaken by means of a questionnaire. This work summarises opinions of experts and findings of said events on the topic of demographic fairness in biometric systems including several important aspects such as the developments of evaluation metrics and standards as well as related issues, e.g. the need for transparency and explainability in biometric systems or legal and ethical issues.   
\end{abstract}


%
\IEEEpeerreviewmaketitle

\section{Introduction}\label{sec:intro}

Biometric technologies \cite{Jain-HandbookBiometrics-2007} have become an integral component of many personal, commercial, and governmental identity management systems worldwide. Biometrics rely on highly distinctive characteristics of human beings, which make it possible for individuals to be reliably recognised using fully automated algorithms. Prominent examples of biometric characteristics used for recognition purposes are face, fingerprint, iris, and voice. Application scenarios of biometrics beyond personal devices (see \eg \cite{Das-MobileBiometrics-2018}) include, but are not limited to border control (see \eg \cite{EULisa-EURODAC-2016,SmartBorders-EU-2018,Northrop-HART-2018}), forensic investigations and law enforcement (see \eg \cite{Moses-AFIS-2010,FBI-CODIS-2018}), and national ID systems (see \eg \cite{UIDAI-Aadhaar-2012}).

According to the international standard ISO/IEC 2382-37 \cite{ISO-Vocabulary-2017}, the term \emph{biometrics} is defined as: “automated recognition of individuals based on their biological and behavioural characteristics”. In an automated biometric system, a capture device (\eg a camera) is used to acquire a biometric sample (\eg facial image) during the enrolment process. Signal processing algorithms are subsequently applied to the biometric sample, which pre-process it (\eg detection and normalisation of the face), estimate the quality of the acquired sample, and extract discriminative features from it. The resulting feature vector is finally stored as reference template. At the time of authentication, another biometric sample is processed in the same way resulting in a probe template. Comparison and decision algorithms enable ascertaining of similarity of a reference and a probe template by comparing the corresponding feature vectors and establish whether or not the two biometric samples belong to the same source.

Automated systems (including biometrics) are increasingly used in decision making processes within various domains. Some of the involved application domains have traditionally enjoyed strong anti-discrimination legislation protection. Such legislations, \eg the anti-discrimination directives of the European Union, aim at protecting against discrimination based on different protected characteristics such as age or race.

\begin{figure}[!ht]
\centering
\setlength{\tabcolsep}{1.5pt}
\def\currentimgsize{0.09}
\renewcommand{\arraystretch}{0.6}
\begin{tabular}{ccccc}
\includegraphics[width=\currentimgsize\textwidth]{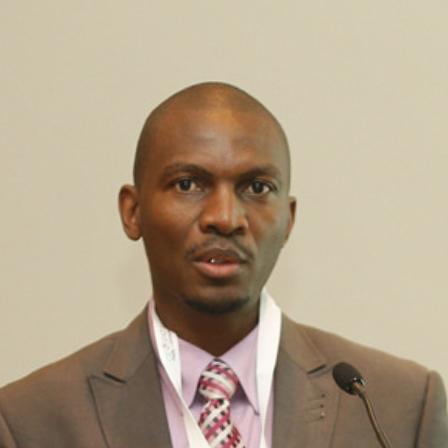} &
\includegraphics[width=\currentimgsize\textwidth]{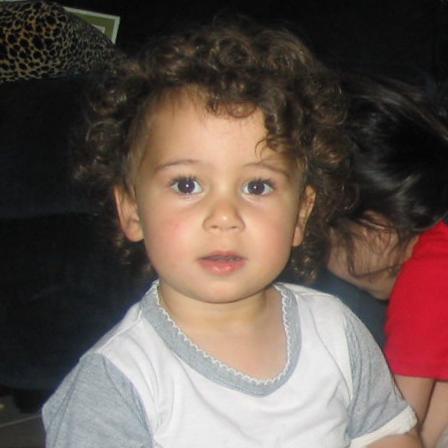} &
\includegraphics[width=\currentimgsize\textwidth]{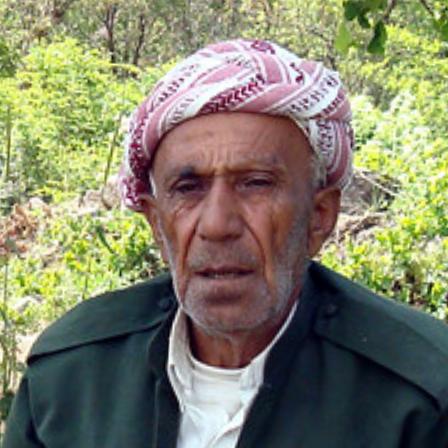} &
\includegraphics[width=\currentimgsize\textwidth]{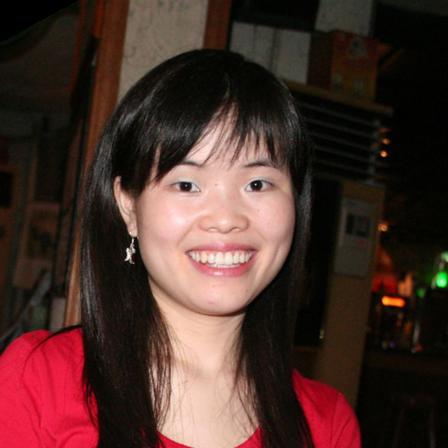} &
\includegraphics[width=\currentimgsize\textwidth]{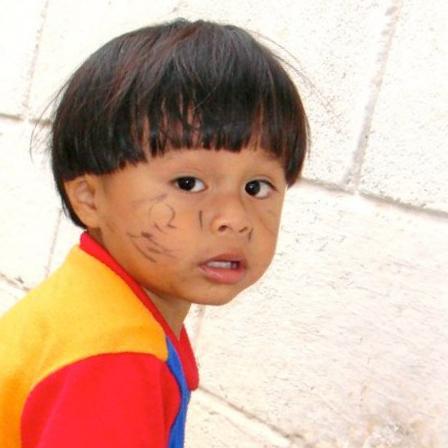} \\
\includegraphics[width=\currentimgsize\textwidth]{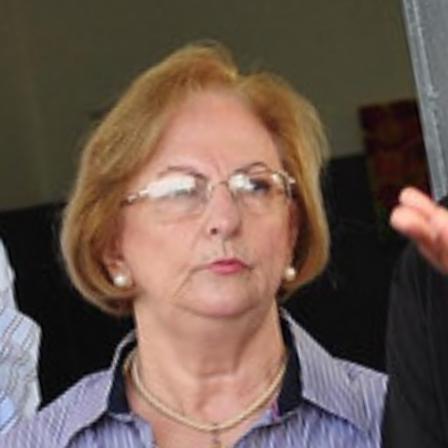} &
\includegraphics[width=\currentimgsize\textwidth]{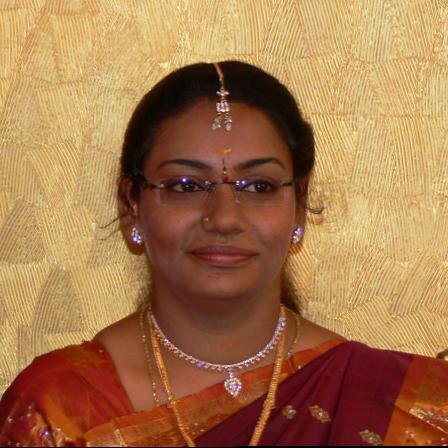} &
\includegraphics[width=\currentimgsize\textwidth]{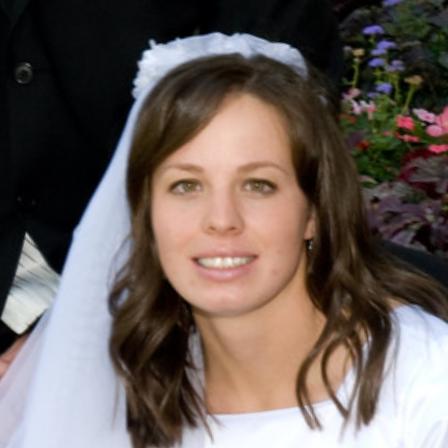} &
\includegraphics[width=\currentimgsize\textwidth]{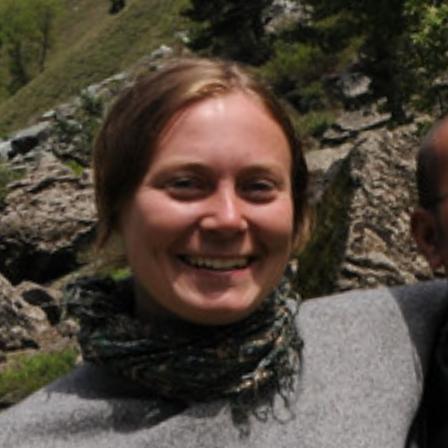} &
\includegraphics[width=\currentimgsize\textwidth]{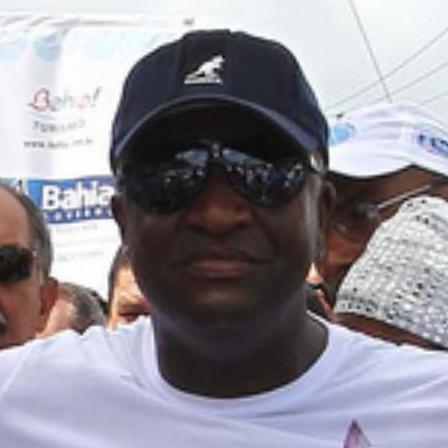} \\
\includegraphics[width=\currentimgsize\textwidth]{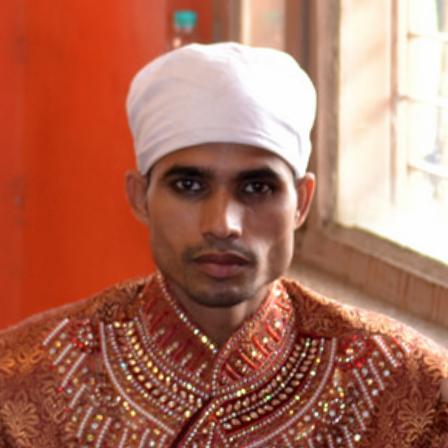} &
\includegraphics[width=\currentimgsize\textwidth]{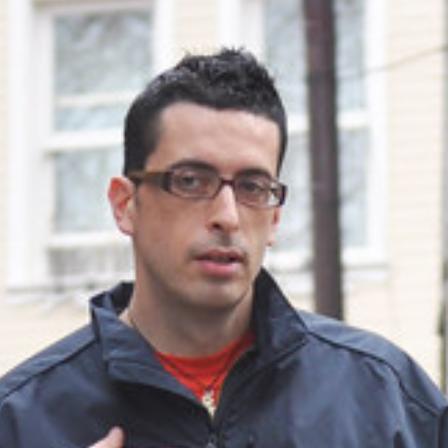} &
\includegraphics[width=\currentimgsize\textwidth]{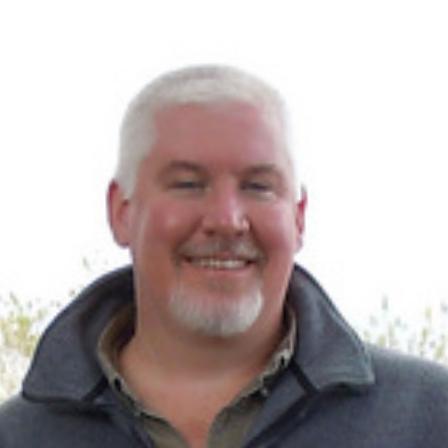} &
\includegraphics[width=\currentimgsize\textwidth]{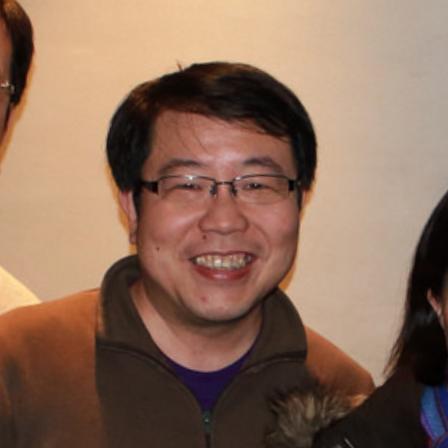} &
\includegraphics[width=\currentimgsize\textwidth]{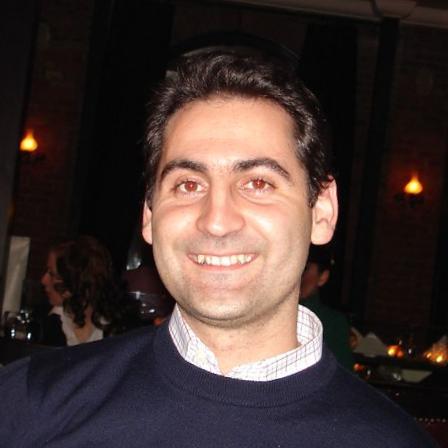} \\
\includegraphics[width=\currentimgsize\textwidth]{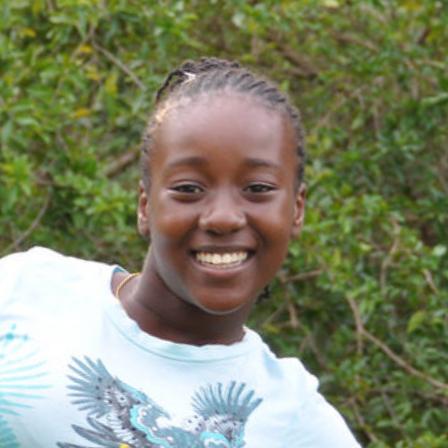} &
\includegraphics[width=\currentimgsize\textwidth]{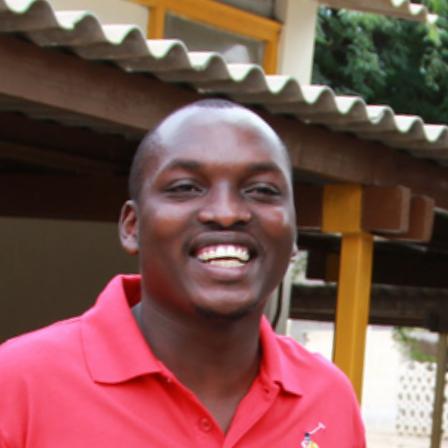} &
\includegraphics[width=\currentimgsize\textwidth]{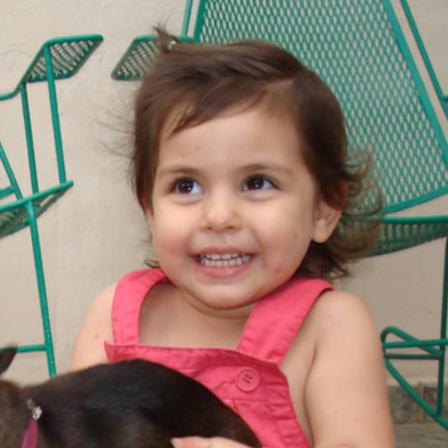} &
\includegraphics[width=\currentimgsize\textwidth]{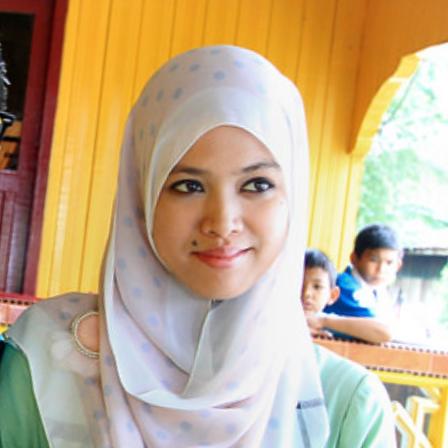} &
\includegraphics[width=\currentimgsize\textwidth]{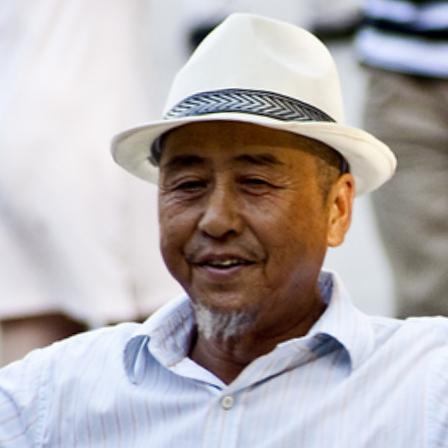} \\
\includegraphics[width=\currentimgsize\textwidth]{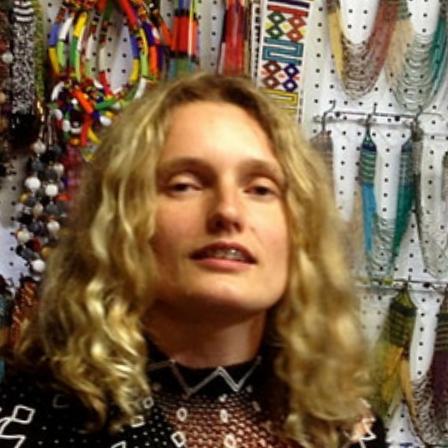} &
\includegraphics[width=\currentimgsize\textwidth]{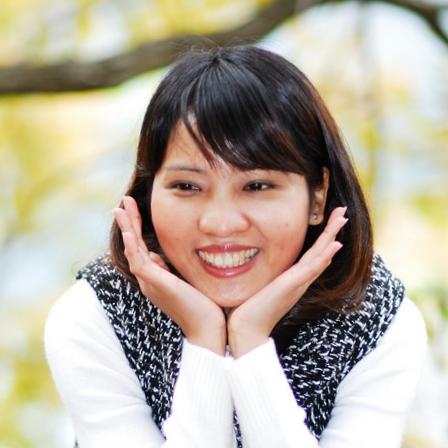} &
\includegraphics[width=\currentimgsize\textwidth]{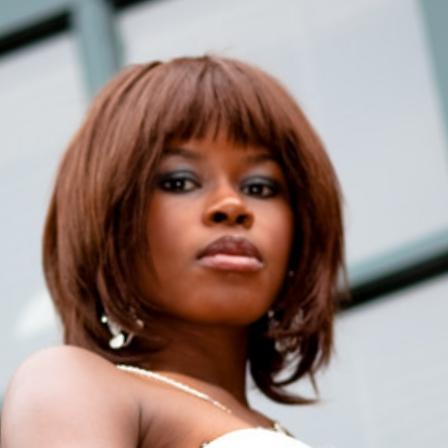} &
\includegraphics[width=\currentimgsize\textwidth]{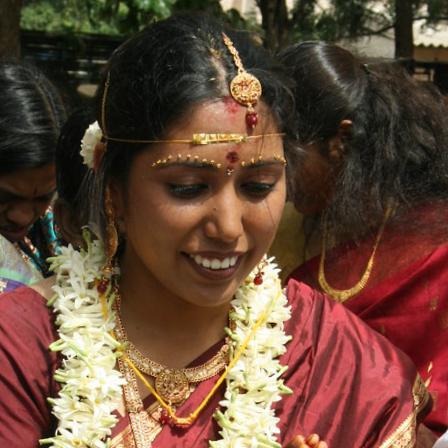} &
\includegraphics[width=\currentimgsize\textwidth]{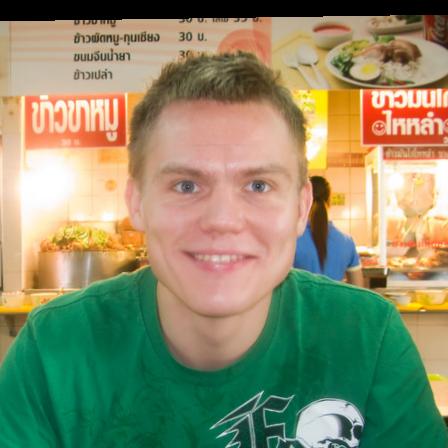} \\
\end{tabular}
\caption{Examples of different demographics which might exhibit performance differentials in a biometric system utilising facial information (images taken from a publicly available research database~\cite{karkkainen2019fairface}).}
\label{fig:demopgraphics}
\end{figure}

In recent years, substantial media coverage of systemic biases inherent to such systems have been reported and hotly debated. In this context, a \emph{biased} algorithm produces statistically different outcomes (decisions) for different groups of individuals, \eg based on sex, age, and race \cite{Nature-Editorial-2016}. In the context of biometric recognition, this means that false positive and/or false negative error rates can differ across the demographic groups. Such \emph{differential outcomes} in biometric systems caused by \emph{demographic performance differentials} can be measured and quantified. In this regard, (demographic) \emph{fairness} is considered a social construct, determined by general agreement and/or laws, which is related to the consequences of differential outcomes, \ie it maps the response/behaviour of a (biometric) algorithm onto an application. The aftermath of demographic differentials can be unsettling -- for example, a large study has reported disproportionally high arrest and search rates of African Americans based on decisions made by automatic facial recognition software \cite{Garvie-PerpetualLineUp-2016}. In fact, the vast majority of works on measuring or achieving fairness in biometric systems is focused on facial recognition \cite{Drozdowski-BiasSurvey-TTS-2020}. Attention has been brought to face since in that biometric modality, performance differentials mostly fall across points of sensitivity (\eg race, sex), see figure~\ref{fig:demopgraphics} for examples of facial images of different demographic groups. Note that the photos may highlight cultural differences in clothing while face recognition algorithms would only process facial regions. Consequently, the overlap between technical and the social issues, \ie adverse effects on classes that have been determined by society (protected classes), makes face recognition an extremely sensitive topic. While most research on demographic fairness is focused on face recognition, biometric systems relying on othe characteristics (in particular behavioural biometrics) still need to be investigated to see whether these might as well exhibit demographic performance differentials.

Obviously, there are other covariates that may negatively influence the performance of biometric recognition systems, \eg environmental factors such as illumination. It is important to note that fairness only relates to differential outcomes correlated with demographics, \ie intrinsic properties of individuals.  However, there might be an overlap between variations in demographics and other, \eg individual, variations. For instance, a beard might be related to an individual style but also to religious beliefs and may, in the latter case, be worn frequently by a certain demographic group \cite{DBLP:journals/corr/abs-2103-01592}. Since biometric systems should work well in every situation, these issues are related. The observation and discussion that biometric systems are not equally fair among (groups of) individuals dates back to Doddigton's zoo menagerie \cite{Doddington-SheepGoatsLambsWolves-1998}, a pioneer work on providing a structure to describe the biometric performance of individuals. In one of the largest evaluations of demographic effects in biometrics, large (orders of magnitude) differential outcomes have been established in numerous commercial algorithms submitted to the NIST Face Recognition Vendor Test (FRVT) benchmark \cite{Grother-NIST-FRVTBias-2019}. In contrast to the face, for some biometric characteristics, \eg fingerprints or finger vein, performance differentials do not necessarily fall across protected groups \cite{Marasco-BiasFingerprint-2019,Drozdowski-BiasFingervein-2020}.

Although biometric systems are supposedly not created to be explicitly unfair against any group, demographic differentials can occur independently of the intentions of the system designers. They can be exhibited (and propagated) at many stages of the decision-making pipeline, including but not limited to training data itself, as well as the data processing \cite{Danks-AlgorithmicBias-IJCAI-2017}. Due to the unprecedented scale and scope of such systems, the potential impact of erroneous or inaccurate decisions may be even higher than in the typical, human-based processes \cite{Tufekci-AlgorithmicHarms-2015}. In recent years, measuring and ensuring the fairness of such systems has often been discussed in the media and the political circles, with the research interest increasing accordingly. Said interestedness notwithstanding, this topic is still relatively new -- comprehensive legal and practical provisions do not yet exist. From the academic point of view, while preliminary studies do exist, the area has by no means been researched exhaustively yet.
Although some studies have approached ensuring fairness in various machine learning contexts (see \eg \cite{Hardt-EqualityLearning-2016}), for computer vision and biometrics in particular, this remains a nascent field with much research potential. 

\begin{figure}[!ht]
\fboxsep=0mm
\fboxrule=1.5pt
\centering
\includegraphics[width=0.5\textwidth]{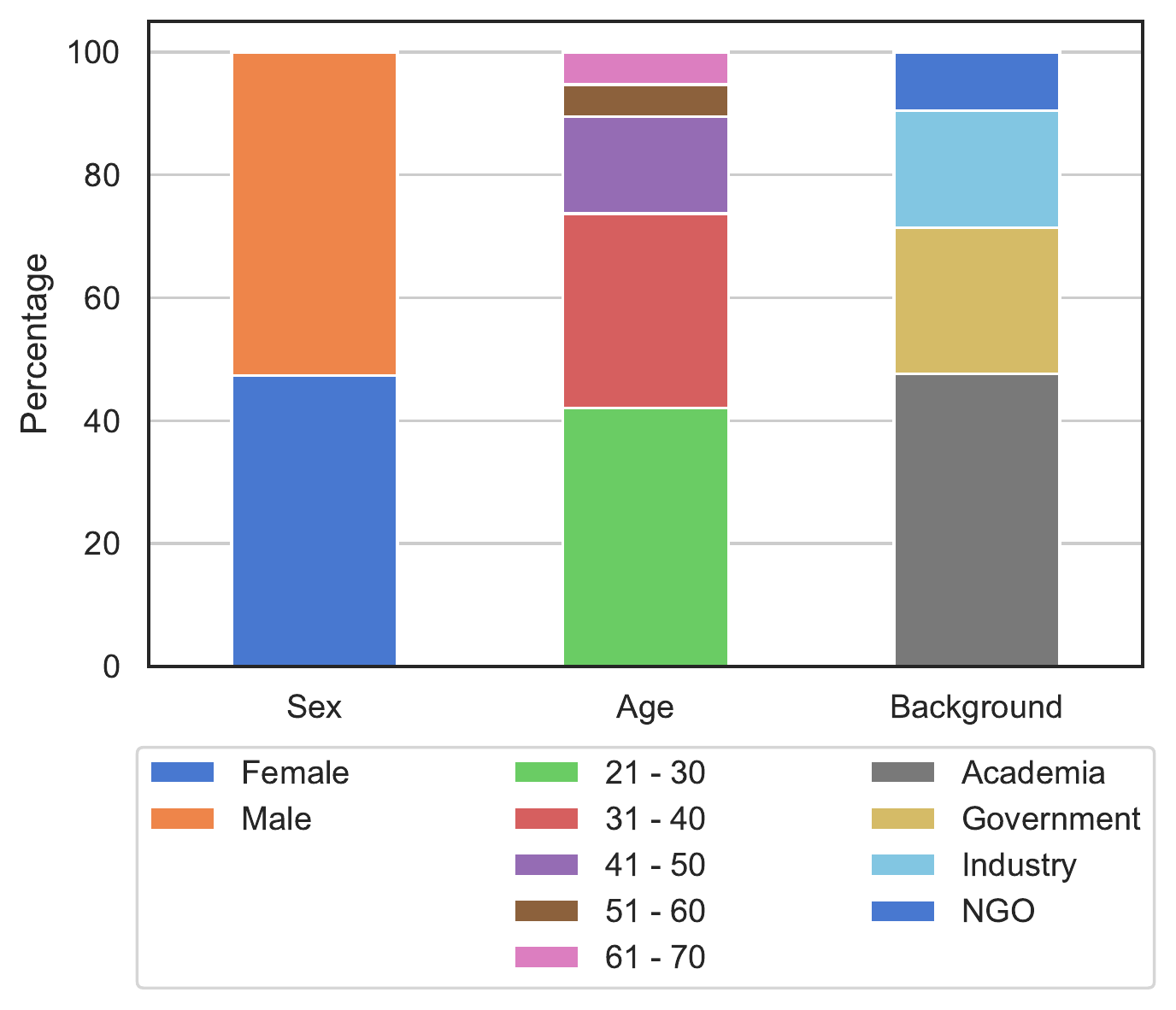}
\caption{Properties of experts participating in the conducted questionaire on demographic fairness in biometrics.}
\label{fig:experts}
\end{figure}

To motivate the research in this field and enhance the problem understanding between stakeholders, in March 2021, the European Association for Biometrics (EAB) \cite{EAB} organised an event series on demographic fairness in biometric systems consisting of two half-day workshops \cite{EABevent} with expert lectures and moderated discussion panels\footnote{EAB is continuously organising events related to research topics in biometrics, see: \url{https://eab.org/events/}}. For these events, more than 120 experts from academia, government, industry, as well as NGOs registered. All registered experts were aware of the debate and research on demographic fairness in biometrics (or more broadly, in algorithmic decision systems) beforehand and were asked to participate in a survey on the topic of demographic fairness in biometric systems. This survey was developed by the organizers prior to the event. It was designed to include only a small list of questions which capture the most important aspects of fairness in biometrics, ranging from technical to legal and societal aspects. The remainder of this this work summarises:

\begin{itemize}
\item The answers of 27 experts who participated in the aforementioned questionaire, see figure~\ref{fig:experts}.
\item Insights gained from the presentations and panel discussions of the aforementioned event series. In this case, discussed findings do not directly correspond to the questionaire.
\item Comments provided by expert reviewers.
\end{itemize}

Collected insights of the above listed sources of information are sumamrised and some quotes of experts are provided. In addition to the summary of the listed sources of information, findings and results reported in the scientific literature are mentioned where appropriate.
This work links the perspectives of different stakeholders on the issue of fairness in biometrics, directs towards future research efforts which are expected to reveal benefits for real-world applications and can adhere to possible future regulations.

The above listed sources of experts' opinions are organised as follows: section~\ref{sec:paradox} discusses techno-social challenges and dilemmas regarding achieving demographic fairness in biometrics and other automated decision systems. Section~\ref{sec:data_priv} stresses the need for privacy-preserving data collections which enable fairness measurement in biometric systems. Important aspects on the definition and standardisation of evaluation metrics are summarised in section~\ref{sec:metrics_stand}. The fundamental need for transparency and explainability as a key requirement for fair and trustworthy biometrics is discussed in section~\ref{sec:trans_explain}. Subsequently, pertinent ethical and legal aspects surrounding the development of fair biometric systems are mentioned in section~\ref{sec:ethical_legal}. Finally, section~\ref{sec:conclusion} concludes this work with additional discussions and future research lines.

\section{The Fairness Dilemma}\label{sec:paradox}

Research in the area of algorithmic fairness concentrates on the following topics:
\begin{itemize}
\item Theoretical and formal definitions of bias and fairness (see \eg~\cite{Mehrabi-BiasSurvey-2019,Verma-FairnessDefinitions-2018,Hutchinson-FairnessSurvey-2019}).
\item Fairness metrics, software, and benchmarks (see \eg~\cite{Hardt-EqualityLearning-2016,Saleiro-BiasToolkit-2018,Bellamy-AIFairness-2018}).
\item Societal, ethical, and legal aspects of algorithmic decision-making and fairness therein (see \eg~\cite{Pasquale-BlackBoxSociety-2015,Tufekci-AlgorithmicHarms-2015,Corbett-AlgorithmicFairness-2017,FRA-BigData-2018,FRA-BiasFundamentalRights-2019}).
\item Estimation and mitigation of bias in algorithms and datasets (see \eg~\cite{Torralba-DatasetBias-2011,Zemel-FairRepresentations-2013,Shaikh-MLFairness-2017,Fernandez-ImbalancedLearning-2018,Zhang-BiasMitigation-2018,Roy-InformationLeakage-2019,Krishnapriya-FaceBiasRace-2020,Albiero-DatasetBalance-2020,TerhorstKDKK20}).
\end{itemize}

Despite several years of research, there exists no single agreed coherent definition of algorithmic fairness. In fact, dozens of formal definitions (see \eg~\cite{Verma-FairnessDefinitions-2018,Hutchinson-FairnessSurvey-2019}) have been proposed to address different situations, which are additionally entangled with different criteria of fairness\footnote{See also \url{https://towardsdatascience.com/a-tutorial-on-fairness-in-machine-learning-3ff8ba1040cb} and \url{https://fairmlbook.org/} for visual tutorials on bias and fairness in machine learning.}. Certain definitions, which are commonly used and advocated for, are even provably mutually exclusive~\cite{Friedler-FairnessImpossibility-2016}.

Fairness is especially critical in algorithms that are explicitly designed around the concept of individual identity, such as biometrics. From a technical point of view, demographic differentials in biometric systems can be measured and quantified. It is generally agreed that one of the key steps for measuring demographic differentials is the collection of adequate datasets. Said datasets are required to contain subsets of a variety of demographic groups and shall also allow for scenario-based evaluations. Conducting such a data collection creates its own challenges (see section~\ref{sec:data_priv}). According to an expert, a fair biometric system should produce the same outcome for different demographics:  \begin{quotation}‘Algorithmic systems should produce the same outcome if all independent variables, except for demographic variables, are the same.’ \end{quotation}

\begin{figure}[!ht]
\fboxsep=0mm
\fboxrule=1.5pt
\centering
\includegraphics[width=0.5\textwidth]{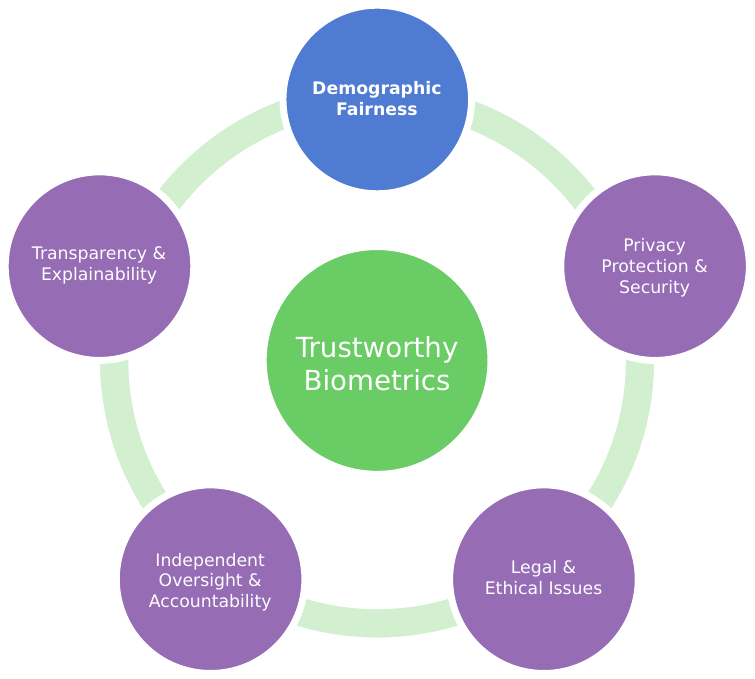}
\caption{Demographic fairness as key factor towards trustworthy biometrics, among others.}
\label{fig:trust}
\end{figure}

 That is, in a fair biometric system, testing on a demographic subset should ideally reveal the same results compared to testing on the entire population. For performing such evaluations first attempts to define suitable metrics have already been made, \eg in \cite{Serna-DeepLearningFaceBias-2019,Freitas-FairBio-2020}, which is considered another essential prerequisite towards reaching fairness in biometric systems. Further, it is of utmost importance to not only compare individual biometric algorithms, but also to put their performance in relation to that of humans. This is the case because the concept of fairness, or the lack of it, also exists in human made decisions \cite{10.1371/journal.pone.0004215}, which can influence both, algorithm design and algorithm evaluation. Ideally, appropriate metrics should be standardised (see section~\ref{sec:metrics_stand}). As mentioned earlier, an unambiguous definition of fairness might only be defined for a specific application scenario. However, it is questionable whether definitions of fairness should solely be based on technical measures. Numerous related issues, \eg transparency and explainability (see section~\ref{sec:trans_explain}) or ethical and legal aspects (see section~\ref{sec:ethical_legal}), are inevitably related with demographic fairness, and need to be solved to achieve trustworthy biometrics, see figure~\ref{fig:trust}. All stated requirements generally agreed upon by experts notwithstanding, a global definition of fairness in biometrics is still lacking. One may argue that fairness needs to be rigorously defined first (possibly in relation to different norms, contexts, and application areas), before it can be improved.

\section{Data Collection and Privacy Protection}\label{sec:data_priv}

\begin{figure}[!ht]
\fboxsep=0mm
\fboxrule=1.5pt
\centering
\includegraphics[width=0.48\textwidth]{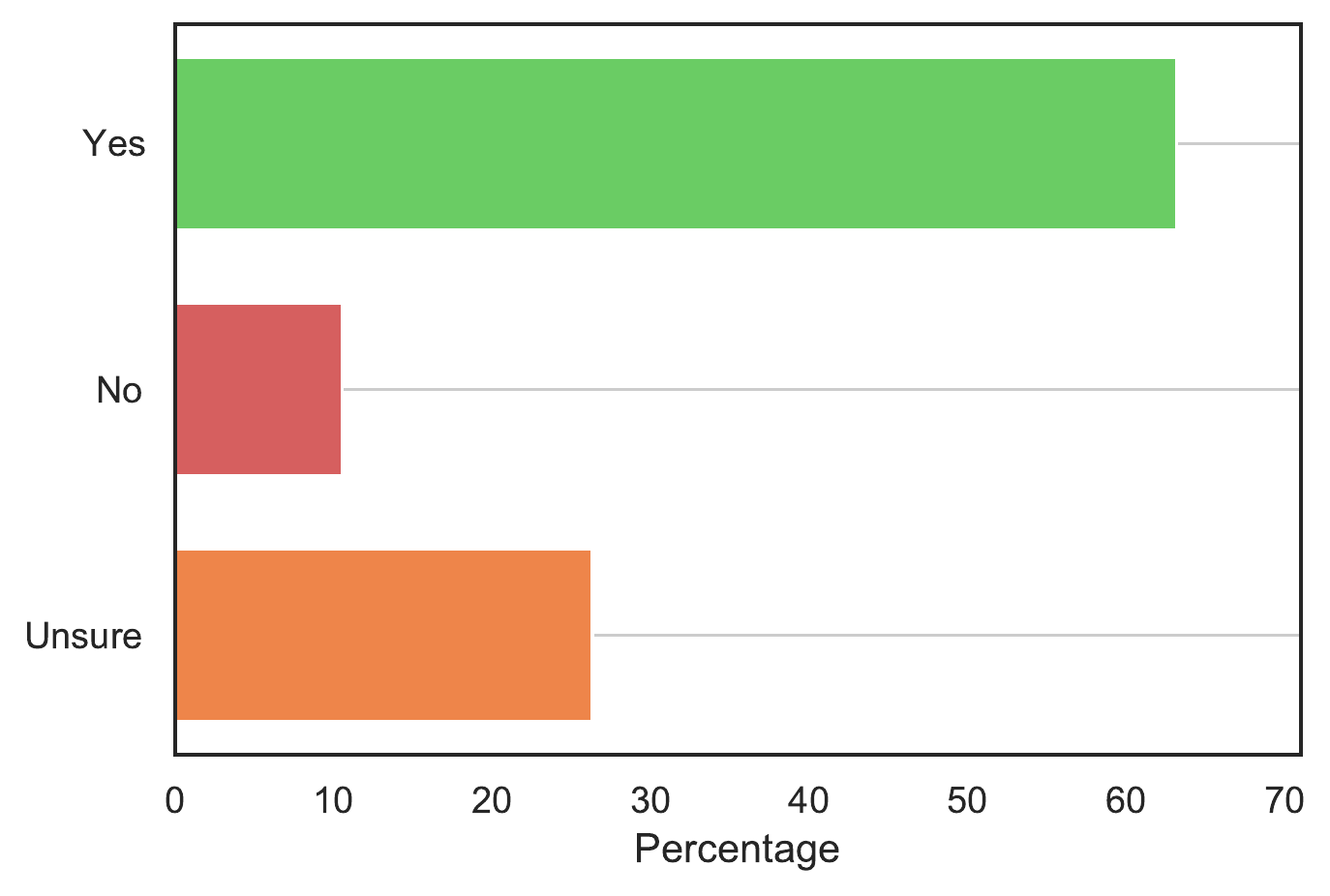}
\caption{Should large-scale, but strongly privacy-conscious, data-collection be initiated to evaluate the demographic differentials in biometric systems?}
\label{fig:collection}
\end{figure}

It is generally agreed among experts that large-scale data collections are vital towards reliable measurement of demographic differentials in biometric recognition systems, see figure~\ref{fig:collection}. Such datasets should not be limited to facial imagery, but ideally comprise a multitude of biometric characteristics including relevant annotations beyond demographics \cite{DBLP:journals/corr/abs-2103-01592}. It is required that the vast majority of demographic groups is represented in such a large-scale database where subsets per country or clusters of countries could be defined. 

The majority of current biometric systems heavily relies on deep learning where massive datasets are employed for algorithm training \cite{Sundararajan-DeepLearningBiometrics-2018}. Research groups have found that a biased training, \ie based on datasets which are unbalanced \wrt demographics, can be the reason behind an unfair biometric system, \eg in \cite{Karkkainen-FairFace-2019}. However, different works have found that a demographically balanced training databases may not be sufficient to improve biometric recognition systems towards making fairer decisions. \cite{Wang_2019_ICCV,Albiero20a}.

Despite the advantages of such a large-scale database, privacy concerns \wrt to the collection of biometric data and subsequent performance trails arise, as stressed by an expert: \begin{quotation}‘Such trials should, however, ensure maximum possible protection of subjects' data, as well as to ensure that the subjects providing their data have full understanding of how their data will be used.’ \end{quotation} In the context of biometric systems, \emph{privacy} relates to the protection of biometric data. Privacy-enhancing technologies that implement fundamental data protection principles have been introduced for protecting stored biometric data which are generally classified as sensitive is considered as highly sensitive in many privacy regulations, \eg the General Data Protection Regulation (GDPR) \cite{EU-GDPR-2016} of the European union.  This is especially true when such a database contains demographic information beyond identity.  Hence, data subjects should be in control of their own data and protection of their privacy must be guaranteed, \eg through biometric template protection \cite{Rathgeb-TemplateProtection-EURASIP-2011,ISO11-TemplateProtection}. Additionally, full transparency on usage must be ensured. \emph{Transparency} in a biometric system means that its processing steps and are easily comprehensible and accessible, \ie subjects providing their data should have full understanding of how their data will be processed/used and they should be able to revoke their consent. With respect to the transparency, various approaches have been recently introduced to enhance the explainability of biometric decisions \cite{Neto2022}. Experts agree that a high extent of transparency and accountability, in particular \wrt privacy and data protection, is needed when conducting such a data collection (see section~\ref{sec:trans_explain}). If a large-scale database should include a huge variety of demographics, it is likely that it has to be collected in different countries. In this regard, it is important to note that there are huge differences in how data privacy is regulated worldwide, if at all \cite{UNCTAD-2020}. An expert summarises, \begin{quotation}‘Transparency of automated decision making systems is a very complex and very prolific research area at the moment and it is not currently clear how to achieve this.’ \end{quotation}

A privacy-compliant way of creating a large-scale biometric database that is balanced regarding demographics can be to fully rely on algorithms that allow for a controllable generation of synthetic biometric data \cite{Joshi22,SFACE}. Different approaches for the creation of synthetic biometric data have been proposed in the scientific literature, \eg for fingerprint \cite{Cappelli-SythteicFinger-2009} or iris \cite{Drozdowski-SICGen-BIOSIG-2017}. Recent advances in deep convolutional neural networks have achieved further improvements in synthetic data generation including biometric data. In particular, Generative Adversarial Networks (GANs) \cite{Goodfellow-GAN-2014}, have shown remarkable results for the generation of biometric data including the face \cite{Karras-SyntehticFace-2020}. The use of such techniques would certainly enable the creation of a demographically balanced synthetic large-scale biometric database, if demographic factors serve as parameters to the synthetisation process. Further, it needs to be assure that generation methods do not suffer from any identity leakage, \ie there should be no correlation between identity infromation contained in the training data and the syntehtically generated data. This means, these methods must not reproduce biometric data that matches the training data which has recently been shown to be the case for some GAN-based generation approaches \cite{Tinsley_2021_WACV}. Whether synthetic biometric data does realistically simulate real world applications, is an question that requires further investigations \cite{Zhang-IWBF-2021}. 

\section{Evaluation Metrics and Standardisation}\label{sec:metrics_stand}

As already mentioned in the preceding sections (see section~\ref{sec:intro} and \ref{sec:paradox}), fairness should be measured \wrt a certain context or application. Therefore, works on metrics to technically measure the fairness of biometric algorithms are frequently limited to estimate performance differentials. It is important to note that in some cases, differential performance may not impact outcomes. More precisely, biometric comparison score distributions may move while being far away from the decision threshold of the biometric system. In this context, experts stress that differential outcomes should be differentiated from differential performances:\begin{quotation}‘It is important to distinguish between differential performance and outcomes to assess actual impact on end-users; in some cases differential performance may not impact outcomes.’ \end{quotation} To assess the actual impact on end-users, differential performance and outcomes need to be distinguished in an application-specific way, \eg impact of being suspected in a watchlist scenario, missing a flight \etc It is important to clearly specify the scenario being tested; the current evaluations may be suitable for technology and scenario evaluations, but some operational conditions may also be highly relevant for tailoring the metrics and evaluation protocols. However, design of an assessment that fits the operational conditions remains a challenge. A feasible path to be followed towards a fairness measure could be to assign weights to the specific scenarios. Depending on the application, differential outcomes according to technical metrics, \eg false non-match rate or false match rate, or operation modes, \ie verification or identification, could be assigned weights to derive an overall fairness measure. However, such an assignment of weights is usually outside of scope of academic investigations and requires joint effort from different stakeholders.

It is generally agreed among experts that complete fairness might not be achievable in biometric systems. For instance, identical performance in terms of false match rate may lead to performance differentials in terms of false non-match rate, and vice versa. Further, demographic fairness may not only depend on the biometric algorithm itself. For instance, if a biometric system is operated in identification mode (one-to-many comparison), performance differentials can result from an uneven composition of demographics in the gallery which may be referred to as \emph{watchlist imbalance effect}. In detail, false matches can occur with significantly higher probabilities within the same demographic group \cite{Howard-DemographicEffectsFace-2019}. Consequently, false matches become more likely in biometric identification trails for demographic groups that are overly represented in a gallery.

The above mentioned issues can partially be tackled by further improving the performance of biometric systems. Experts agree that demographic fairness and biometric performance are highly correlated. A general reduction of error rates would decrease the absolute number of differential outcomes in a biometric system and, hence, improve fairness, as argued by an expert: \begin{quotation}‘As algorithms become more accurate there are fewer errors for all demographic groups, so more accurate algorithms help.’ \end{quotation}
 In contrast, current works report a trade-off between demographic differentials and recognition accuracy, \ie a decrease of performance differentials frequently yields a decrease of the overall biometric performance \cite{Drozdowski-BiasSurvey-TTS-2020}. In some biometric systems, a mitigation of demographic performance differentials may be achieved by forcefully ignoring some attributes, \eg sex, in order to improve demographic fairness which obviously reduces the amount of extracted information. Considering the inevitable limitations of some biometric systems, a reduction of error rates to a level where performance differentials become insignificant is considered challenging. Experts agree that good biometric sample quality \cite{ISO-Quality-2016} is a prerequisite for performance improvements in biometric systems. Finally, it is important to establish what constitutes ``unfair'', especially at very low overall error rates. For this purpose, certain (application-specific) criteria need to be introduced, similarly to existing measures for other use-cases, \eg the four-fifths rule used by the uniform guidelines on employee selection procedures \cite{Bobko04}.

\begin{figure}[!ht]
\fboxsep=0mm
\fboxrule=1.5pt
\centering
\includegraphics[width=0.45\textwidth]{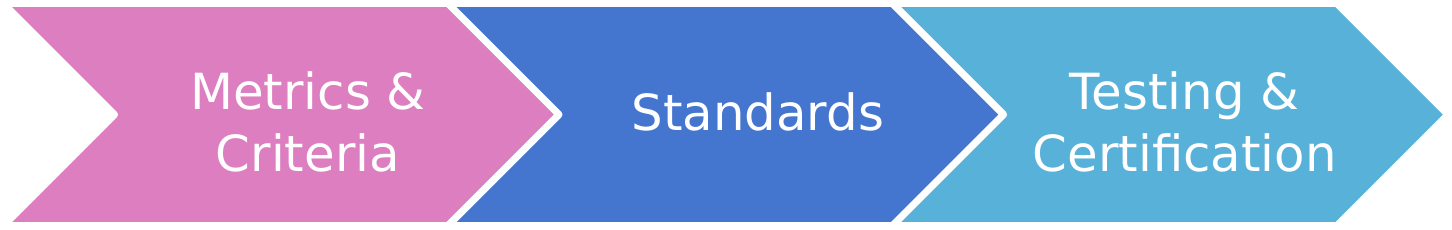}
\caption{Metrics to measure fairness in biometric systems need to be defined in standards which are required for testing and certification.}
\label{fig:certification}
\end{figure}

For the goal of certifying biometric systems \wrt demographic fairness, standards need to be established which comprise evaluation metrics and criteria to be applied, see figure~\ref{fig:certification}. Note that an international group of experts is currently working on a dedicated ISO/IEC standard \cite{ISO-Bias}, which is considered as an important activity: \begin{quotation}‘Independent bias performance evaluation or even certification could be a step forward.’ \end{quotation} The use of such standardised metrics and evaluation methodologies further enables a transparent comparison of biometric systems and sets a precondition for deployment. Moreover, experts suggest that suitable databases might be distributed together with standards. Independent benchmarks conceptually similar to \eg NIST FRVT \cite{Grother-NIST-FRVTBias-2019} could also be considered in the context of systems' certification. It is further suggested that said standards may also include measures to evaluate human fairness, \eg for manual border control. It has been shown that humans' face recognition abilities reveal demographic performance differentials as well, in particular the other-race effect has been prominently documented, \eg \cite{Phillips-FaceRace-2011}. In this regard, it is important to note that experts share the opinion that some automated biometric systems are likely to be more fair than humans. On the one hand, this would be something that needs to be better communicated; on the other hand, this suggests that human fairness may not represent a suitable criteria to decide whether fairness is fulfilled in a biometric system. In addition, it is concluded that besides biometric algorithms, standardised evaluation metrics should be used to test human performance to identify training needs.

\section{Transparency, Explainability, and Oversight}\label{sec:trans_explain}

The aforementioned efforts on standardisation are expected to lead to more transparency and, hence, acceptability of biometric systems, as mentioned by an expert:\begin{quotation}‘Standardization and documentation of biometric algorithm will lead to more transparency and transparency will lead to acceptability.’\end{quotation} Beyond that, individuals who use biometric algorithms should be aware of further properties of the system, \eg how data is stored, \etc (ideally, biometric data should stay in the hands of the user). Furthermore, it is necessary for users to understand the decision process of a biometric system. On the one hand, experts believe that a more transparent decision-making process is a key requirement towards achieving trustworthy biometrics. This is a major issue which is related to explaining what is learned and encoded within neural network systems based on a specific training dataset. On the other hand, it is important for operators in order to adjust the biometric system's decision if required, \ie by inserting a human in the loop \cite{Jain-Trust-2021}.  

Regarding the use of machine learning, it is agreed among experts that explainability represents a challenging problem in many state-of-the-art biometric systems. However, it is also important to note that experts perceive a disconnect between what biometric system developers are trying to accomplish and what the public perceives it is being used for. Consequently, there is a great need to communicate the principle of operation of biometric systems and their impact to the general public and how such technologies will protect, not harm, their everyday lives. Furthermore, it is noteworthy that many automated decision-making systems which are not based on machine learning may also lack explainability and, more importantly, human decisions may not be transparent, explainable, or as accurate either.

\begin{figure}[!ht]
\fboxsep=0mm
\fboxrule=1.5pt
\centering
\includegraphics[width=0.48\textwidth]{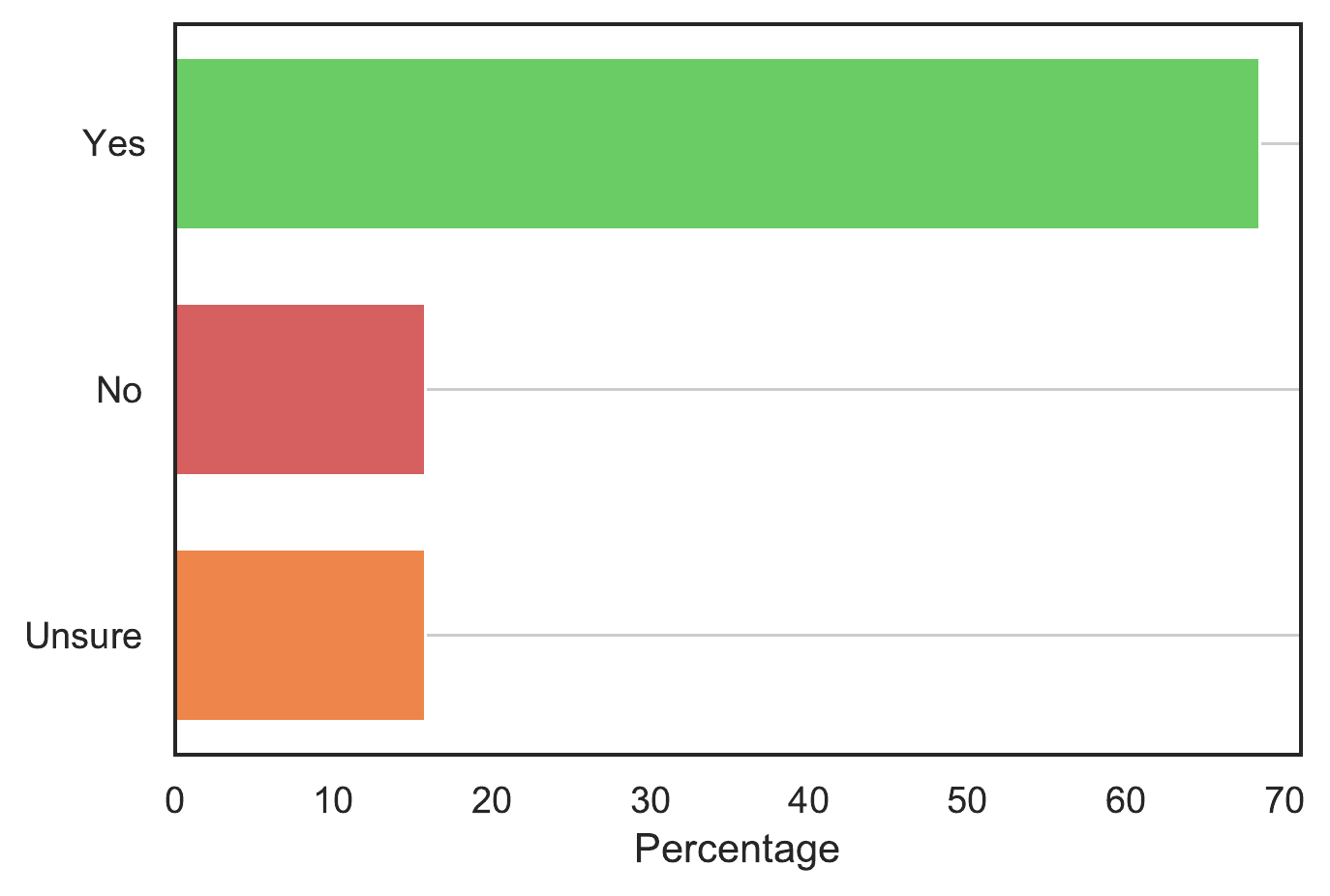}
\caption{Should independent oversight, in principle, be considered for deployments of algorithmic decision systems, including biometrics? (e.g. surveillance, border control).}
\label{fig:oversight}
\end{figure}

According to many experts in this field, independent oversight should be considered for deployments of biometric systems, see figure~\ref{fig:oversight}. The size and scope of such a system as well as a differentiation between commercial and governmental operators could constitute some of the viable criteria to decide on whether such oversight is required. Moreover, the application scenario and its impact should be considered, \eg automated border access control applications are different from watchlist identifications.

\section{Ethical and Legal Aspects}\label{sec:ethical_legal}

Ethical and legal aspects play an important role towards developing fair biometric systems. Regarding the collection of biometric databases, privacy regulations need to be taken into account (see section~\ref{sec:data_priv}). For instance, in the European Union the GDPR \cite{EU-GDPR-2016}, regulates the use of biometric data. On the one hand, the GDPR is considered as good from a user's point of view, since it imposes obligations on data controllers. On the other hand, this privacy regulation is challenging from the operator's point of view, since it frames its use. This means, laws regarding the use of biometric data may improve the users' trust in biometric technologies while they may slow down the development of fairer biometric systems. As previously mentioned, some experts identify the use of synthetic biometric data as one solution to this problem.  

It is agreed among experts that legal rules should be based on use-cases and their impact, \eg what degree of harm could be caused. Both commercial and government operators should be required to fulfill the same obligations. Lawmakers and regulators already have experiences in regulating fairness in other fields, \eg discrepancies in hiring rates or payment. However, experts warn that even if rules/measurements are stipulated for some use-cases, these usually tend to be highly ambiguous and, hence, mostly shift burdens. Furthermore, there is currently no measure that could be directly re-used for biometric systems. Communication to lawmakers and regulators to create awareness of the issue of demographic fairness in biometric systems is considered to be another important issue. Moreover, experts recommend that oversight by legally independent authorities should be mandatory for (large-scale) biometric systems (see section~\ref{sec:trans_explain}).

From an ethical point of view, some experts warn against the discrimination of certain groups \wrt the use of biometric systems, \eg individuals with disabilities, defects, and other issues that prevent them from using one or more of the most widely used biometric technologies. Further, experts identify the lack of socio-economic awareness or educational background of operators of biometric systems, \eg border guards, as potential source for ethical issues.

\section{Conclusion}\label{sec:conclusion}

Quantifying demographic differentials in biometric systems and identifying sources thereof enables informed policy decisions by the relevant stakeholders, while simultaneously being fundamental for the development of unbiased, fair algorithms in the future. Those matters are directly related to the deployment of biometric systems, in particular ensuring uniformly high quality of service and usability for all the system users. Furthermore, demographic performance differentials can directly affect the biometric recognition performance, and hence negatively impact the usability and security of the systems. The measurement and mitigation of demographic performance differentials in biometric systems still represents a nascent field of research \cite{Drozdowski-BiasSurvey-TTS-2020}. 

This work summarised opinions of experts on the topic of demographic fairness in biometric systems as well as other related topics. Experts' inputs were obtained in the course of an event series organised by the EAB. In summary, the following key recommendations towards fair biometric systems can be derived from the experts' opinions: 
\begin{itemize}
\item Large-scale biometric databases need to be collected in order to reliably measure demographic performance differentials. Collections shall be performed in a transparent and privacy compliant manner. In this context, synthetic biometric data generation should be investigated as an alternative privacy compliant solution.
\item Standardized evaluation metrics shall be developed and applied in order to measure and compare demographic differentials in a transparent manner. Here, differential performance and outcome needs to be distinguished and the application scenario should be taken into account.
\item Towards demographic fairness, the explainability of biometric decisions requires improvement. At the same time, independent oversight, in particular by legal and ethics experts, should be considered for deployments of algorithmic decision systems including biometrics.
 \end{itemize}

\section*{Acknowledgements}

The authors would like to thank all the experts who contributed to this work and the reviewers for valuable comments and discussions. 

This work was partially funded by the German Federal Ministry of Education and Research and the Hessian Ministry of Higher Education, Research, Science and the Arts within their joint support of the National Research Center for Applied Cybersecurity ATHENE.

\ifCLASSOPTIONcaptionsoff
  \newpage
\fi

\bibliographystyle{IEEEtran}
\bibliography{references}

\begin{thebibliography}{10}
\providecommand{\url}[1]{#1}
\csname url@samestyle\endcsname
\providecommand{\newblock}{\relax}
\providecommand{\bibinfo}[2]{#2}
\providecommand{\BIBentrySTDinterwordspacing}{\spaceskip=0pt\relax}
\providecommand{\BIBentryALTinterwordstretchfactor}{4}
\providecommand{\BIBentryALTinterwordspacing}{\spaceskip=\fontdimen2\font plus
\BIBentryALTinterwordstretchfactor\fontdimen3\font minus
  \fontdimen4\font\relax}
\providecommand{\BIBforeignlanguage}[2]{{%
\expandafter\ifx\csname l@#1\endcsname\relax
\typeout{** WARNING: IEEEtran.bst: No hyphenation pattern has been}%
\typeout{** loaded for the language `#1'. Using the pattern for}%
\typeout{** the default language instead.}%
\else
\language=\csname l@#1\endcsname
\fi
#2}}
\providecommand{\BIBdecl}{\relax}
\BIBdecl

\bibitem{Jain-HandbookBiometrics-2007}
A.~K. Jain, P.~Flynn, and A.~Ross, \emph{Handbook of biometrics}.\hskip 1em
  plus 0.5em minus 0.4em\relax Springer, 2007.

\bibitem{Das-MobileBiometrics-2018}
A.~Das, C.~Galdi, H.~Han, R.~Ramachandra, J.-L. Dugelay, and A.~Dantcheva,
  ``Recent advances in biometric technology for mobile devices,'' in
  \emph{International Conference on Biometrics Theory, Applications and Systems
  ({BTAS})}.\hskip 1em plus 0.5em minus 0.4em\relax IEEE, October 2018, pp.
  1--11.

\bibitem{EULisa-EURODAC-2016}
S.~European Union Agency for the Operational Management of Large-Scale IT
  Systems in the Area~of Freedom and Justice, ``Eurodac storage capacity
  increased,''
  \url{https://www.eulisa.europa.eu/Newsroom/News/Pages/Eurodac-storage-capacity-increased.aspx},
  April 2016, last accessed: \today.

\bibitem{SmartBorders-EU-2018}
{European Commission}, ``Smart borders,''
  \url{https://ec.europa.eu/home-affairs/what-we-do/policies/borders-and-visas/smart-borders_en},
  2018, last accessed: \today.

\bibitem{Northrop-HART-2018}
T.~Paynter, ``{Northrop Grumman} wins \$95 million award from {Department of
  Homeland Security} to develop next-generation biometric identification
  services system,''
  \url{https://news.northropgrumman.com/news/releases/northrop-grumman-wins-95-million-award-from-department-of-homeland-security-to-develop-next-generation-biometric-identification-services-system},
  February 2018, last accessed: \today.

\bibitem{Moses-AFIS-2010}
K.~R. Moses, P.~Higgins, M.~McCabe, S.~Probhakar, and S.~Swann,
  \emph{Fingerprint Sourcebook}.\hskip 1em plus 0.5em minus 0.4em\relax US
  Department of Justice, 2010, ch. Automated Fingerprint Identification System
  ({AFIS}), pp. 1--33.

\bibitem{FBI-CODIS-2018}
F.~B. of~Investigation, ``{CODIS} - {NDIS} statistics,''
  \url{https://www.fbi.gov/services/laboratory/biometric-analysis/codis/ndis-statistics},
  June 2018, last accessed: \today.

\bibitem{UIDAI-Aadhaar-2012}
{Unique Identification Authority of India}, ``Role of biometric technology in
  {A}adhaar enrollment,'' {UIDAI}, Tech. Rep., January 2012.

\bibitem{ISO-Vocabulary-2017}
{ISO/IEC JTC1 SC37 Biometrics}, \emph{{ISO/IEC} 2382-37:2017. Information
  technology -- Vocabulary -- Part~37: Biometrics}, 2nd~ed., International
  Organization for Standardization and International Electrotechnical
  Committee, February 2017.

\bibitem{Nature-Editorial-2016}
Editorial, ``Algorithm and blues,'' \emph{Nature}, pp. 1--1, September 2016.

\bibitem{Garvie-PerpetualLineUp-2016}
C.~Garvie, \emph{The perpetual line-up: Unregulated police face recognition in
  {A}merica}.\hskip 1em plus 0.5em minus 0.4em\relax Georgetown Law, Center on
  Privacy \& Technology, October 2016.

\bibitem{Drozdowski-BiasSurvey-TTS-2020}
P.~Drozdowski, C.~Rathgeb, A.~Dantcheva, N.~Damer, and C.~Busch, ``Demographic
  bias in biometrics: A survey on an emerging challenge,'' \emph{Transactions
  on Technology and Society ({TTS})}, vol.~1, no.~2, pp. 89--103, June 2020.

\bibitem{karkkainen2019fairface}
K.~K{\"a}rkk{\"a}inen and J.~Joo, ``{FairFace}: Face attribute dataset for
  balanced race, gender, and age,'' \emph{arXiv preprint arXiv:1908.04913},
  2019.

\bibitem{DBLP:journals/corr/abs-2103-01592}
P.~Terh{\"{o}}rst, J.~N. Kolf, M.~Huber, F.~Kirchbuchner, N.~Damer, A.~Morales,
  J.~Fi{\'{e}}rrez, and A.~Kuijper, ``A comprehensive study on face recognition
  biases beyond demographics,'' \emph{arXiv preprint arXiv:2103.01592}, March
  2021.

\bibitem{Doddington-SheepGoatsLambsWolves-1998}
G.~Doddington, W.~Liggett, A.~Martin, M.~Przybocki, and D.~Reynolds, ``Sheep,
  goats, lambs and wolves: A statistical analysis of speaker performance in the
  {NIST} 1998 speaker recognition evaluation,'' in \emph{International
  Conference on Spoken Language Processing}.\hskip 1em plus 0.5em minus
  0.4em\relax Australian Speech Science and Technology Association, December
  1998, pp. 1351--1355.

\bibitem{Grother-NIST-FRVTBias-2019}
P.~Grother, M.~Ngan, and K.~Hanaoka, ``Ongoing face recognition vendor test
  ({FRVT}) part 3: Demographic effects,'' National Institute of Standards and
  Technology, Tech. Rep. NISTIR 8280, December 2019.

\bibitem{Marasco-BiasFingerprint-2019}
E.~Marasco, ``Biases in fingerprint recognition systems: Where are we at?'' in
  \emph{International Conference on Biometrics: Theory Applications and Systems
  ({BTAS})}.\hskip 1em plus 0.5em minus 0.4em\relax IEEE, September 2019, pp.
  1--5.

\bibitem{Drozdowski-BiasFingervein-2020}
P.~Drozdowski, B.~Prommegger, G.~Wimmer, R.~Schraml, C.~Rathgeb, A.~Uhl, and
  C.~Busch, ``Demographic bias: A challenge for fingervein recognition
  systems?'' in \emph{European Signal Processing Conf. ({EUSIPCO})}.\hskip 1em
  plus 0.5em minus 0.4em\relax EURASIP, August 2020, pp. 825--829.

\bibitem{Danks-AlgorithmicBias-IJCAI-2017}
D.~Danks and A.~J. London, ``Algorithmic bias in autonomous systems,'' in
  \emph{International Joint Conference on Artificial Intelligence
  ({IJCAI})}.\hskip 1em plus 0.5em minus 0.4em\relax IJCAI, August 2017, pp.
  4691--4697.

\bibitem{Tufekci-AlgorithmicHarms-2015}
Z.~Tufekci, ``Algorithmic harms beyond {F}acebook and {G}oogle: Emergent
  challenges of computational agency,'' \emph{Journal on Telecommunications and
  High Technology Law}, vol.~13, 2015.

\bibitem{Hardt-EqualityLearning-2016}
M.~Hardt, E.~Price, N.~Srebro \emph{et~al.}, ``Equality of opportunity in
  supervised learning,'' in \emph{Advances in Neural Information Processing
  Systems ({NIPS})}.\hskip 1em plus 0.5em minus 0.4em\relax Neural Information
  Processing Systems Foundation, December 2016, pp. 3315--3323.

\bibitem{EAB}
EAB, ``European association for biometrics,'' \url{https://eab.org/}, last
  accessed: \today.

\bibitem{EABevent}
E.~A. for Biometrics, ``Eab virtual events series – demographic fairness in
  biometric systems,'' \url{https://eab.org/events/program/237}, last accessed:
  \today.

\bibitem{Mehrabi-BiasSurvey-2019}
N.~Mehrabi, F.~Morstatter, N.~Saxena, K.~Lerman, and A.~Galstyan, ``A survey on
  bias and fairness in machine learning,'' \emph{arXiv preprint
  arXiv:1908.09635}, September 2019.

\bibitem{Verma-FairnessDefinitions-2018}
S.~Verma and J.~Rubin, ``Fairness definitions explained,'' in
  \emph{International Workshop on Software Fairness ({FairWare})}.\hskip 1em
  plus 0.5em minus 0.4em\relax ACM, May 2018, pp. 1--7.

\bibitem{Hutchinson-FairnessSurvey-2019}
B.~Hutchinson and M.~Mitchell, ``50 years of test (un)fairness: Lessons for
  machine learning,'' in \emph{Conference on Fairness, Accountability, and
  Transparency ({FAT})}.\hskip 1em plus 0.5em minus 0.4em\relax ACM, January
  2019, pp. 49--58.

\bibitem{Saleiro-BiasToolkit-2018}
P.~Saleiro, B.~Kuester, A.~Stevens, A.~Anisfeld, L.~Hinkson, J.~London, and
  R.~Ghani, ``Aequitas: A bias and fairness audit toolkit,'' \emph{arXiv
  preprint arXiv:1811.05577}, November 2018.

\bibitem{Bellamy-AIFairness-2018}
R.~K.~E. Bellamy, K.~Dey, M.~Hind, S.~C. Hoffman, S.~Houde \emph{et~al.},
  ``{AI} fairness 360: An extensible toolkit for detecting, understanding, and
  mitigating unwanted algorithmic bias,'' \emph{arXiv preprint
  arXiv:1810.01943}, October 2018.

\bibitem{Pasquale-BlackBoxSociety-2015}
F.~Pasquale, \emph{The black box society: The secret algorithms that control
  money and information}.\hskip 1em plus 0.5em minus 0.4em\relax Harvard
  University Press, 2015.

\bibitem{Corbett-AlgorithmicFairness-2017}
S.~Corbett-Davies, E.~Pierson, A.~Feller, S.~Goel, and A.~Huq, ``Algorithmic
  decision making and the cost of fairness,'' in \emph{International Conference
  on Knowledge Discovery and Data Mining}.\hskip 1em plus 0.5em minus
  0.4em\relax ACM, August 2017, pp. 797--806.

\bibitem{FRA-BigData-2018}
F.~Focus, ``Bigdata: Discrimination in data-supported decision making,''
  European Union Agency for Fundamental Rights, Tech. Rep. TK-02-18-634-EN-N,
  May 2018.

\bibitem{FRA-BiasFundamentalRights-2019}
F.~Focus, ``Data quality and artificial intelligence -- mitigating bias and
  error to protect fundamental rights,'' European Union Agency for Fundamental
  Rights, Tech. Rep. TK-01-19-330-EN-N, June 2019.

\bibitem{Torralba-DatasetBias-2011}
A.~Torralba and A.~A. Efros, ``Unbiased look at dataset bias,'' in
  \emph{Conference on Computer Vision and Pattern Recognition ({CVPR})}.\hskip
  1em plus 0.5em minus 0.4em\relax IEEE, June 2011, pp. 1521--1528.

\bibitem{Zemel-FairRepresentations-2013}
R.~Zemel, Y.~Wu, K.~Swersky, T.~Pitassi, and C.~Dwork, ``Learning fair
  representations,'' in \emph{International Conference on Machine Learning
  ({ICML})}.\hskip 1em plus 0.5em minus 0.4em\relax JMLR, February 2013, pp.
  325--333.

\bibitem{Shaikh-MLFairness-2017}
S.~Shaikh, H.~Vishwakarma, S.~Mehta, K.~R. Varshney, K.~N. Ramamurthy, and
  D.~Wei, ``An end-to-end machine learning pipeline that ensures fairness
  policies,'' \emph{arXiv preprint arXiv:1710.06876}, October 2017.

\bibitem{Fernandez-ImbalancedLearning-2018}
A.~Fern{\'a}ndez, S.~Garc{\'i}a, M.~Galar, R.~C. Prati, B.~Krawczyk, and
  F.~Herrera, \emph{Learning from Imbalanced Data Sets}.\hskip 1em plus 0.5em
  minus 0.4em\relax Springer, October 2018.

\bibitem{Zhang-BiasMitigation-2018}
B.~H. Zhang, B.~Lemoine, and M.~Mitchell, ``Mitigating unwanted biases with
  adversarial learning,'' in \emph{Conference on AI, Ethics, and Society
  ({AIES})}.\hskip 1em plus 0.5em minus 0.4em\relax ACM, February 2018, pp.
  335--340.

\bibitem{Roy-InformationLeakage-2019}
P.~C. Roy and V.~N. Boddeti, ``Mitigating information leakage in image
  representations: A maximum entropy approach,'' in \emph{Conference on
  Computer Vision and Pattern Recognition ({CVPR})}.\hskip 1em plus 0.5em minus
  0.4em\relax IEEE, June 2019, pp. 2586--2594.

\bibitem{Krishnapriya-FaceBiasRace-2020}
K.~S. Krishnapriya, V.~Albiero, K.~Vangara, M.~C. King, and K.~W. Bowyer,
  ``Issues related to face recognition accuracy varying based on race and skin
  tone,'' \emph{Transactions on Technology and Society ({TTS})}, vol.~1, no.~1,
  pp. 8--20, March 2020.

\bibitem{Albiero-DatasetBalance-2020}
V.~Albiero, K.~Zhang, Bowyer, and K.~W. Bowyer, ``How does gender balance in
  training data affect face recognition accuracy?'' in \emph{International
  Joint Conference on Biometrics ({IJCB})}.\hskip 1em plus 0.5em minus
  0.4em\relax IEEE, September 2020, pp. 1--10.

\bibitem{TerhorstKDKK20}
P.~Terh{\"{o}}rst, J.~N. Kolf, N.~Damer, F.~Kirchbuchner, and A.~Kuijper,
  ``Post-comparison mitigation of demographic bias in face recognition using
  fair score normalization,'' \emph{Pattern Recognit. Lett.}, vol. 140, pp.
  332--338, 2020.

\bibitem{Friedler-FairnessImpossibility-2016}
S.~A. Friedler, C.~Scheidegger, and S.~Venkatasubramanian, ``On the
  (im)possibility of fairness,'' \emph{arXiv preprint arXiv:1609.07236},
  September 2016.

\bibitem{Serna-DeepLearningFaceBias-2019}
I.~Serna, A.~Morales, J.~Fierrez, M.~Cebrian, N.~Obradovich, and I.~Rahwan,
  ``Algorithmic discrimination: Formulation and exploration in deep
  learning-based face biometrics,'' \emph{arXiv preprint arXiv:1912.01842},
  December 2019.

\bibitem{Freitas-FairBio-2020}
T.~de~Freitas~Pereira and S.~Marcel, ``Fairness in biometrics: a figure of
  merit to assess biometric verification systems,'' \emph{arXiv preprint
  arXiv:2011.02395}, November 2020.

\bibitem{10.1371/journal.pone.0004215}
S.~Lebrecht, L.~J. Pierce, M.~J. Tarr, and J.~W. Tanaka, ``Perceptual
  other-race training reduces implicit racial bias,'' \emph{PLOS ONE}, vol.~4,
  no.~1, pp. 1--7, January 2009.

\bibitem{Sundararajan-DeepLearningBiometrics-2018}
K.~Sundararajan and D.~L. Woodard, ``Deep learning for biometrics: a survey,''
  \emph{Computing Surveys ({CSUR})}, vol.~51, no.~3, pp. 65:1--65:34, July
  2018.

\bibitem{Karkkainen-FairFace-2019}
K.~K{\"a}rkk{\"a}inen and J.~Joo, ``{FairFace}: Face attribute dataset for
  balanced race, gender, and age,'' \emph{arXiv preprint arXiv:1908.04913},
  August 2019.

\bibitem{Wang_2019_ICCV}
T.~Wang, J.~Zhao, M.~Yatskar, K.-W. Chang, and V.~Ordonez, ``Balanced datasets
  are not enough: Estimating and mitigating gender bias in deep image
  representations,'' in \emph{International Conference on Computer Vision
  ({ICCV})}, 2019.

\bibitem{Albiero20a}
V.~Albiero, K.~Zhang, and K.~W. Bowyer, ``How does gender balance in training
  data affect face recognition accuracy?'' in \emph{Int'l Joint Conference on
  Biometrics (IJCB)}, 2020, pp. 1--10.

\bibitem{EU-GDPR-2016}
{European Parliament}, ``Regulation (eu) 2016/679,'' \emph{Official Journal of
  the {European Union}}, vol. L119, pp. 1--88, April 2016.

\bibitem{Rathgeb-TemplateProtection-EURASIP-2011}
C.~Rathgeb and A.~Uhl, ``A survey on biometric cryptosystems and cancelable
  biometrics,'' \emph{{EURASIP} Journal on Information Security}, 2011.

\bibitem{ISO11-TemplateProtection}
{ISO/IEC JTC 1/SC 27 IT Security techniques}, \emph{{ISO/IEC} 24745:2011.
  Information technology -- Security techniques -- Biometric information
  protection}, International Organization for Standardization and International
  Electrotechnical Committee, June 2011.

\bibitem{Neto2022}
P.~C. Neto, T.~Gonçalves, J.~R. Pinto, W.~Silva, A.~F. Sequeira, A.~Ross, and
  J.~S. Cardoso, ``Explainable biometrics in the age of deep learning,''
  \emph{arXiv preprint arXiv:2208.09500}, 2022.

\bibitem{UNCTAD-2020}
{United Nations Conference on Trade and Development}, ``Data protection and
  privacy legislation worldwide,''
  \url{https://unctad.org/page/data-protection-and-privacy-legislation-worldwide},
  April 2020, last accessed: \today.

\bibitem{Joshi22}
I.~Joshi, M.~Grimmer, C.~Rathgeb, C.~Busch, F.~Bremond, and A.~Dantcheva,
  ``Synthetic data in human analysis: A survey,'' \emph{arXiv preprint
  arXiv:2208.09191}, 2022.

\bibitem{SFACE}
F.~Boutros, M.~Huber, P.~Siebke, T.~Rieber, and N.~Damer, ``{SFace}:
  Privacy-friendly and accurate face recognition using synthetic data,'' in
  \emph{International Joint Conference on Biometrics ({IJCB})}, 2022.

\bibitem{Cappelli-SythteicFinger-2009}
R.~Cappelli, \emph{Synthetic Fingerprint Generation}.\hskip 1em plus 0.5em
  minus 0.4em\relax Springer London, 2009, pp. 271--302.

\bibitem{Drozdowski-SICGen-BIOSIG-2017}
P.~Drozdowski, C.~Rathgeb, and C.~Busch, ``{SIC-Gen}: {A} synthetic {Iris-Code}
  generator,'' in \emph{Intl. Conf. of the Biometrics Special Interest Group
  ({BIOSIG})}, September 2017, pp. 1--6.

\bibitem{Goodfellow-GAN-2014}
I.~Goodfellow, J.~Pouget-Abadie, M.~Mirza, B.~Xu, D.~Warde-Farley, S.~Ozair,
  A.~Courville, and Y.~Bengio, ``Generative adversarial nets,'' in
  \emph{Advances in Neural Information Processing Systems ({NIPS})},
  vol.~27.\hskip 1em plus 0.5em minus 0.4em\relax Curran Associates, Inc.,
  2014.

\bibitem{Karras-SyntehticFace-2020}
T.~Karras, S.~Laine, and T.~Aila, ``A style-based generator architecture for
  generative adversarial networks,'' \emph{Transactions on Pattern Analysis and
  Machine Intelligence}, pp. 1--1, January 2020.

\bibitem{Tinsley_2021_WACV}
P.~Tinsley, A.~Czajka, and P.~Flynn, ``This face does not exist... but it might
  be yours! identity leakage in generative models,'' in \emph{Winter Conference
  on Applications of Computer Vision (WACV)}, 2021, pp. 1320--1328.

\bibitem{Zhang-IWBF-2021}
H.~Zhang, M.~Grimmer, R.~Raghavendra, K.Raja, and C.Busch, ``On the
  applicability of synthetic data for face recognition,'' in \emph{Intl.
  Workshop on Biometrics and Forensics ({IWBF})}.\hskip 1em plus 0.5em minus
  0.4em\relax IEEE, May 2021, pp. 1--6.

\bibitem{Howard-DemographicEffectsFace-2019}
J.~J. Howard, Y.~B. Sirotin, and A.~R. Vemury, ``The effect of broad and
  specific demographic homogeneity on the imposter distributions and false
  match rates in face recognition algorithm performance,'' in
  \emph{International Conference on Biometric, Theory, Applications and Systems
  ({BTAS})}.\hskip 1em plus 0.5em minus 0.4em\relax IEEE, September 2019.

\bibitem{ISO-Quality-2016}
{ISO/IEC JTC1 SC37 Biometrics}, \emph{{ISO/IEC} 29794-1:2016. Information
  technology -- Biometric sample quality -- Part~1: Framework}, International
  Organization for Standardization and International Electrotechnical
  Committee, September 2016.

\bibitem{Bobko04}
P.~Bobko and P.~Roth, ``The four-fifths rule for assessing adverse impact: An
  arithmetic, intuitive, and logical analysis of the rule and implications for
  future research and practice,'' \emph{Research in Personnel and Human
  Resources Management}, vol.~23, pp. 177--198, 2004.

\bibitem{ISO-Bias}
{ISO/IEC JTC1 SC37 Biometrics}, ``{ISO/IEC} {WD} 19795-10 information
  technology -- biometric performance testing and reporting -- part 10:
  Quantifying biometric system performance variation across demographic
  groups,'' unpublished draft.

\bibitem{Phillips-FaceRace-2011}
P.~J. Phillips, F.~Jiang, A.~Narvekar, J.~Ayyad, and A.~J. O'Toole, ``An
  other-race effect for face recognition algorithms,'' \emph{Transactions on
  Applied Perception ({TAP})}, vol.~8, no.~2, pp. 14:1--14:11, January 2011.

\bibitem{Jain-Trust-2021}
A.~K. Jain, D.~Deb, and J.~J. Engelsma, ``Biometrics: Trust, but verify,''
  \emph{arXiv preprint arXiv:2105.06625}, May 2021.

\end{thebibliography}

\newpage
\appendix[Questionnaire]

\begin{enumerate}
\item What is your professional background?
    \begin{enumerate}[label=(\alph*)]
    \item Government
    \item NGO
    \item Industry
    \item Academia
    \item Other (free text)
    \end{enumerate}
\item Which age group are you in?
    \begin{enumerate}[label=(\alph*)]
    \item 21 - 30
    \item 31 - 40
    \item 41 - 50
    \item 51 - 60
    \item 61 - 70
    \item None of the above (free text)
    \end{enumerate}
\item What is your gender?
    \begin{enumerate}[label=(\alph*)]
    \item Female
    \item Male
    \item None of the above (free text)
    \end{enumerate}
\item What are your expectations for this series of events?\\Free text
\item Prior to learning about this event, were you aware of the debate and research on demographic fairness in biometrics, or more broadly, in algorithmic decision systems? 
    \begin{enumerate}[label=(\alph*)]
    \item No
    \item Yes
    \begin{enumerate}[label=(\roman*)]
    		\item What do you believe to be the most important challenges and open questions in this context?\\Free text
    		\item How should biometric fairness w.r.t. demographic factors be defined and measured?\\Free text
    		\item Should large-scale, but strongly privacy-conscious, data-collection be initiated to evaluate the demographic differentials in biometric systems?\\Free text
	\end{enumerate}  
    \end{enumerate}    
\item Should independent oversight be, in principle, considered for deployments (e.g. surveillance or border control) of algorithmic decision systems, including biometrics? 
    \begin{enumerate}[label=(\alph*)]
    \item No
    \item Yes
    \begin{enumerate}[label=(\roman*)]
    		\item What criteria should be used to decide this? For example, based on the size and scope of such a system, or differentiating between commercial and governmental operators, \etc\\Free text
	\end{enumerate}  
    \end{enumerate}        
    \item What extent of transparency, accountability, and explainability should be expected from deployments (e.g. governmental) of biometric technologies?\\Free text
    \item Beyond algorithmic fairness, what do you consider to be the current largest challenges and open questions in biometrics?\\Free text
\end{enumerate}

\end{document}